\UseRawInputEncoding  %
\documentclass[letterpaper]{article} 
\usepackage[]{aaai25}  
\usepackage{times}  
\usepackage{helvet}  
\usepackage{courier}  
\usepackage[hyphens]{url}  
\usepackage{graphicx} 
\usepackage{natbib}  
\usepackage{caption} 
\usepackage{algorithm}
\usepackage{algorithmic}
\usepackage{amsmath}
\usepackage{amssymb} 
\usepackage{xcolor}
\usepackage{booktabs}
\usepackage{multirow}
\usepackage{rotating}
\usepackage{adjustbox}
\usepackage{afterpage}

\usepackage{newfloat}
\usepackage{listings}
\DeclareCaptionStyle{ruled}{labelfont=normalfont,labelsep=colon,strut=off} 
\floatstyle{ruled}
\newfloat{listing}{tb}{lst}{}
\floatname{listing}{Listing}

\title{Exploring Model Editing for LLM-based Aspect-Based Sentiment Classification}

\author{
    Shichen Li\textsuperscript{\rm 1},
    Zhongqing Wang\textsuperscript{\rm 1}\thanks{Corresponding author.},
    Zheyu Zhao\textsuperscript{\rm 1},
    Yue Zhang\textsuperscript{\rm 2},
    Peifeng Li\textsuperscript{\rm 1}
}

\affiliations{
    \textsuperscript{\rm 1}Natural Language Processing Lab, Soochow University, Suzhou, China\\
    \textsuperscript{\rm 2}Westlake University \\
    \{scli06, zyzhao0104\}@stu.suda.edu.cn, 
    \{zqwang, pfli\}@suda.edu.cn \\
    zhangyue@westlake.edu.cn
}

\usepackage{bibentry}

\begin{document}

\maketitle

\begin{abstract}
Model editing aims at selectively updating a small subset of a neural model's parameters with an interpretable strategy to achieve desired modifications.
It can significantly reduce computational costs to adapt to large language models (LLMs).
Given its ability to precisely target critical components within LLMs, model editing shows great potential for efficient fine-tuning applications. 
In this work, we investigate model editing to serve an efficient method for adapting LLMs to solve aspect-based sentiment classification.
Through causal interventions, we trace and determine which neuron hidden states are essential for the prediction of the model. By performing interventions and restorations on each component of an LLM, we identify the importance of these components for aspect-based sentiment classification.
Our findings reveal that a distinct set of mid-layer representations is essential for detecting the sentiment polarity of given aspect words. 
Leveraging these insights, we develop a model editing approach that focuses exclusively on these critical parts of the LLM, leading to a more efficient method for adapting LLMs. 
Our in-domain and out-of-domain experiments demonstrate that this approach achieves competitive results compared to the currently strongest methods with significantly fewer trainable parameters, highlighting a more efficient and interpretable fine-tuning strategy.
\end{abstract}

\section{Introduction}
\begin{figure}[t]
    \captionsetup{type=figure}
    \centering
    \includegraphics[width=\linewidth]{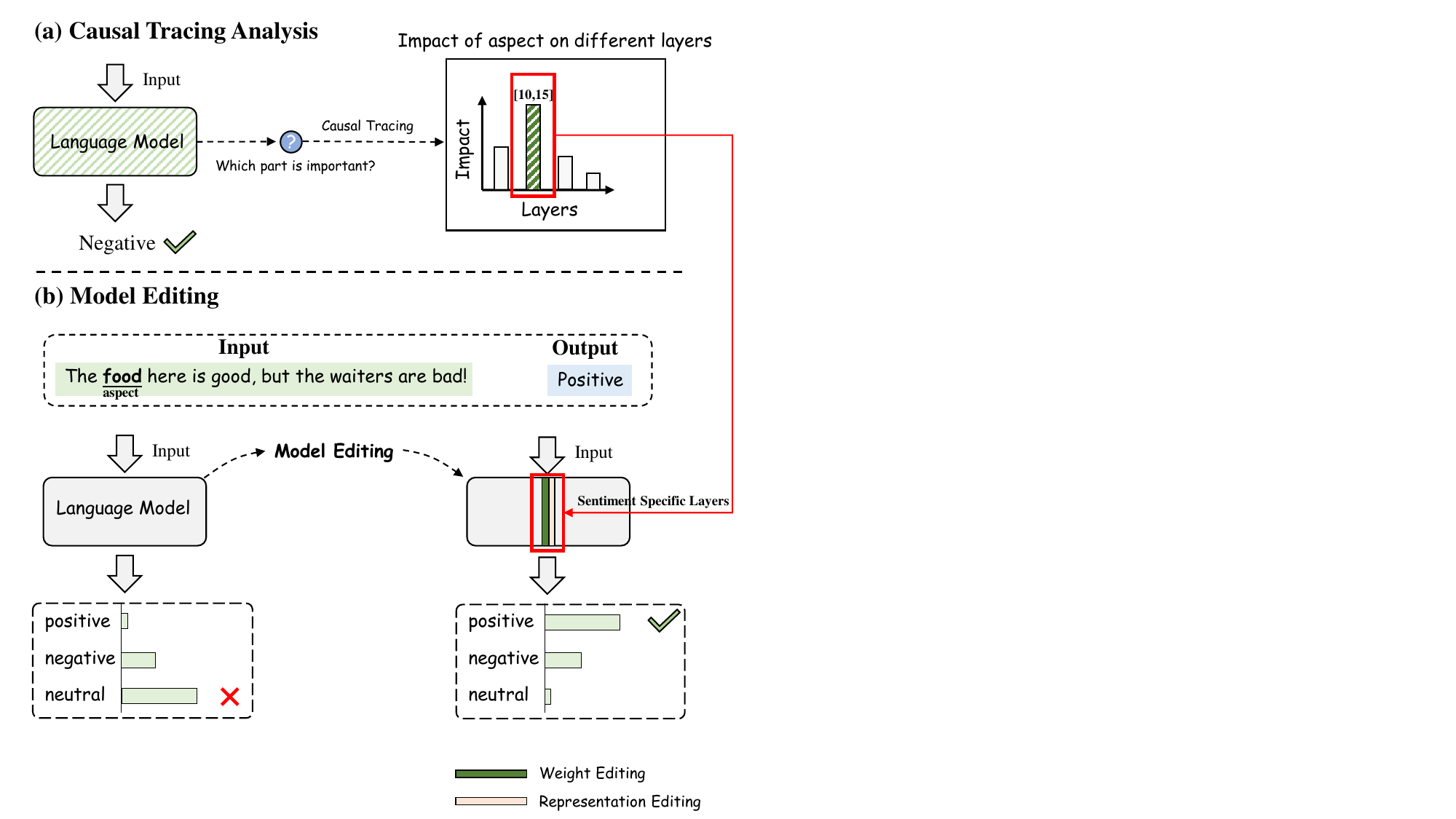}
    \caption{
        \label{fig:intro} 
             An illustration of our 2-steps model editing method for LLM-based aspect-based sentiment classification.
    }
\end{figure}
Aspect-based sentiment classification~\cite{HuL04} aims to determine the sentiment polarity of the aspects within a sentence.
Traditional ABSC methods~\cite{TangQFL16,TayTH18a,ZhangZW22} have often relied on smaller, task-specific language models which frequently fall short in capturing complex contextual dependencies and diverse expressions of sentiment.

As language models continue to grow in size, adapting these large language models~\cite{touvron2023llama,chatgpt22} to ABSC through full-parameter fine-tuning has become increasingly challenging.
Various parameter-efficient fine-tuning methods~\cite{hu2021lora,Adapter19,li2021prefix,liudora2024} have been developed to address the inefficiencies of full-parameter fine-tuning and shown strong performance in ABSC~\cite{liSC22}, but these methods still need to update across broad sections of the model's architecture, potentially leading to redundancies and inefficiencies. 

Model editing~\cite{modeleiding2022,meng2022locating,dai-etal-2022-knowledge} gives a promising way to further reduce the cost by selectively updating a small subset of a neural model's parameters with an interpretable strategy and achieve desired modifications.

Recent works in model editing~\cite{mengMEMIT,modeleiding2022,meng2022locating, wu2024reft} has shown effectiveness and efficiency in various tasks by modifying LLMs with minimal changes to their parameters. 
These methods highlight that targeted updates to specific neural network components can achieve precise modifications, thereby avoiding the extensive resource requirements of full-parameter fine-tuning. 
However, the research question of how to edit LLMs for ABSC is still under-explored.

Intuitively, aspect words significantly influence the model's output, allowing us to perform precise causal analysis and examine how changes to specific components affect overall performance. 
This enables us to accurately identify and modify the model parameters closely associated with sentiment classification, thereby validating the potential of model editing in enhancing model adaptability and efficiency. 

Given the above observations, we consider a model editing method to improve parameter-efficient fine-tuning for ABSC.
The method consists of two steps as shown in Figure~\ref{fig:intro}.
As shown in Figure~\ref{fig:intro}(a), we first employ causal interventions to identify which components are critical for the LLM’s predictions associated with sentiment.
By performing interventions and restorations on each layer of the model, we determine their relative importance for the ABSC task. 
Experiments reveal that the mid-layer representations of specific aspect words in certain positions are particularly critical for accurately identifying the sentiment polarity. 
Based on the insights, we propose a precise model editing method that targets critical parts of the LLM using two model editing approaches: weight-based and representation-based editing, as shown in Figure~\ref{fig:intro}(b). 
For weight-based editing~\cite{hu2021lora}, we use low-rank adaptation to edit specific model weights. 
For representation-based editing~\cite{wu2024reft}, we manipulate a small fraction of model representations at specific positions of aspect terms to steer the model's behavior. 

We conduct extensive experiments to evaluate the method in both in-domain and out-of-domain scenarios, aiming to validate its effectiveness and generalization.
The experiments confirm that our method achieves competitive results with a minimal number of parameters in both scenarios. 
Additionally, we empirically investigate the impact of the proposed model editing method on specific layers and analyze the influence of modifications at specific positions of different words. 
The results show that the mid-layer representations of specific aspect words are crucial for accurately obtaining the correct predictions, aligning with the findings of causal tracing results.

The main contributions of this work can be summarized as follows:
  \begin{itemize}
  	\item We propose a novel framework that combines causal tracing and model editing to address specific downstream tasks with LLMs. 
  	\item We develop a more parameter-efficient fine-tuning method specifically tailored for the aspect-based sentiment classification task. 
  	\item The proposed method achieves competitive results compared to state-of-the-art PEFT methods with significantly fewer trainable parameters, highlighting a more efficient and interpretable fine-tuning strategy.
  \end{itemize}

\section{Related Work}
In this section, we introduce two related topics of this study: aspect-based sentiment classification and model editing.
\subsection{Aspect-Based Sentiment Classification}
Aspect-based sentiment classification methods have seen significant changes with the development of neural networks~\cite{zhang2023survey}. 
Early approaches often relied on full-parameter fine-tuning of smaller models such as LSTM and CNN with attention mechanisms to solve ABSC task~\cite{TangQFL16,MaLZW17,HuangC18,LiX18}. However, these models had limitations in understanding complex contexts and long-range dependencies.
The development of pre-trained language models brought significant  improvement~\cite{SunHQ19,XuLSY19,JiangHZYWP20}. 
By fine-tuning these models on ABSC tasks, researchers were able to better capture the context and improve ABSC performance. 
However, as models grew larger, the computational cost and resource demands of full-parameter fine-tuning became increasingly prohibitive. 
To address these challenges, parameter-efficient fine-tuning methods~\cite{hu2021lora,liudora2024} emerged. 
These methods enable the adaptation of LLMs to ABSC tasks by tuning a smaller subset of parameters. 
Due to the strong foundational capabilities of LLMs, parameter-efficient fine-tuning methods can achieve competitive performance compared to previous carefully designed methods~\cite{liSC22}.
While these methods only update a small portion of parameters within each layer, the updates still span the entire model, leading to some unnecessary redundancies.

\subsection{Model Editing}
With the development of large language models, full-parameters fine-tuning has become challenging. 
To alleviate this inefficiency, model editing~\cite{meng2022locating,mengMEMIT} is designed to update these models in a highly targeted and precise manner. 
Recent advancements in model editing~\cite{yao-etal-2023-editing} can be categorized into two types: knowledge integration and parameter modification.

Knowledge integration aims to augment the model with new information without updating its original parameters. Recent works on activation steering~\cite{steer-embedding,turner2023activation,avitan2024changed} and representation engineering~\cite{zou2023representation,liu2023context} shows that adding fixed or task-specific steering vectors to the model enable precise control over outputs of LMs. 

Parameter modification focuses on directly adjusting the model's parameters. Recent works focus on updating specific parts of the model responsible for particular knowledge. 
Techniques like Knowledge Neurons~\cite{dai-etal-2022-knowledge} and ROME~\cite{meng2022locating} identify and edit specific neurons or layers with a designed causal tracing approach. 
These work make an assumption that the feed-forward network stores knowledge in a key-value format, allowing for directly targeted parameter adjustments to update or add memories.
Another line of work~\cite{wu2024reft} focuses on training interventions to directly edit model representation during inference for specific tasks.

\begin{figure*}[t]
    \captionsetup{type=figure}
    \centering
    \includegraphics[width=\linewidth]{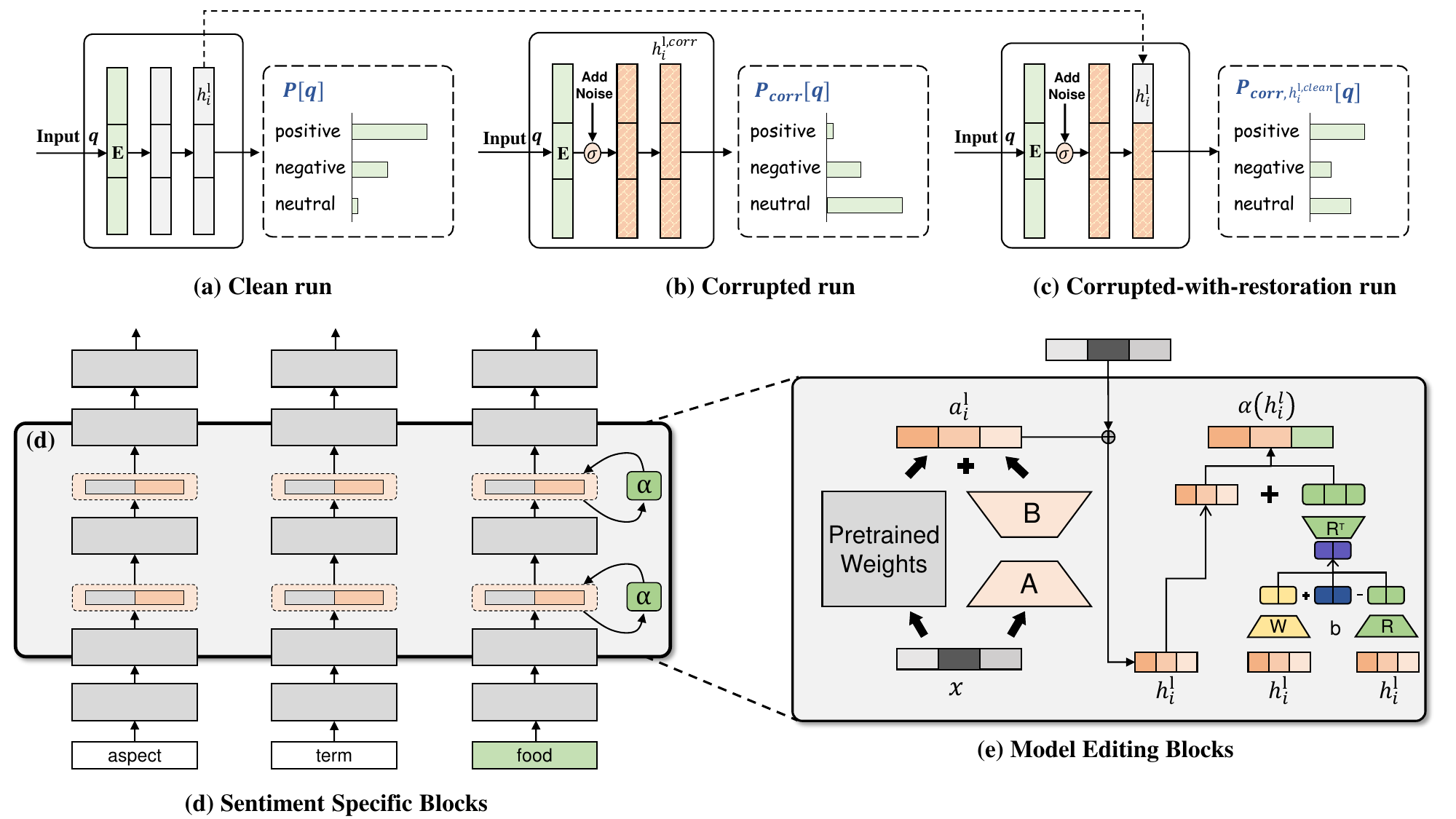}
    \caption{
        \label{fig:flow_chart} 
           Illustration of the process of causal tracing and model editing for ABSC. (a), (b), (c) illustrate three runs for tracing sentiment associations; (d) and (e) demonstrate our method, where colored blocks represent active parameters and the grey blocks represent frozen parameters. Notably, representation-based editing is  applied only to the position of the aspect word.
    }
\end{figure*}
\section{Method}

In this section, we first introduce the intervention method for tracing sentiment associations, then analyze the causal tracing results. Finally, based on the insights of the analysis, we explore model editing as an efficient adaptation method using task-specific model edits.

\subsection{Interventions for Tracing Sentiment Association} 
We begin by identifying the specific layer with the strongest causal effect on sentiment polarity predictions of the given aspect terms by the method proposed in ~\cite{meng2022locating}. 
This process is inspired by the causal tracing technique in ~\cite{vig2020investigating,pearl2022direct,geiger2021causal}, which demonstrates how modifications to hidden states can impact the network's output. 
By systematically adding noise and restoring representations, we can evaluate the sensitivity of the model's predictions to changes in certain components.

We present each sample as a tuple $t = (S, A, P)$ containing the sentence $S$, the aspect $A$, and the sentiment polarity $P$ according to $A$. 
Then, we provide a natural language prompt $q$ describing $(S, A)$ and examine the model's prediction of $q$. 
To identify the layers with the strongest causal effect, we perform interventions on the output of each layer. Specifically, we introduce the following three runs:
\paragraph{Clean run.}
 We input the sentence $S$ and aspect $A$ described by $q$ into the model, and observe the prediction $P$, as shown in Figure~\ref{fig:flow_chart}(a). 
 All hidden representations $\{h_i^{(l)}\}_{i \in [1, T], l \in [1, L]}$ are recorded, where $T$ is the number of tokens and $L$ is the number of layers. 
 
\paragraph{Corrupted run.}
As shown in Figure~\ref{fig:flow_chart}(b), we add noise $\sigma$ to the hidden states at embedding layers denoted as $\{h_i^{(0)}+\sigma\}_{i \in [1, T]}$. 
It results in corrupted representations $\{h_i^{(l),\text{corr}}\}_{i \in [1, T], l \in [1, L]}$. This noise follows the prior practice~\cite{meng2022locating}. The corrupted representations lead to a potentially incorrect sentiment polarity prediction $P^{\text{corr}}$. 

\paragraph{Corrupted-with-restoration run.}
As shown in Figure~\ref{fig:flow_chart}(c), we selectively restore the clean hidden representation $h_i^{(l), \text{clean}}$ at certain token and layer while keeping the rest of the representation corrupted. This run tests if the model can correctly predict the sentiment polarity $P$ despite the overall corruption, indicating the causal importance of the restored hidden states.

Let $\mathbb{P}[q]$, $\mathbb{P}_{\text{corr}}[q]$, and $\mathbb{P}_{\text{corr}, h_i^{(l), \text{clean}}}[q]$ denote the probabilities of predicting $P$ under the clean, corrupted, and corrupted-with-restoration runs, respectively. The causal effect is quantified as follows:

The total effect (TE) is defined as the difference in prediction probabilities between the clean and corrupted runs:
\begin{equation}
\text{TE} = \mathbb{P}[q] - \mathbb{P}_{\text{corr}}[q]
\end{equation}

The indirect effect (IE) is defined as the difference between the probability of correct prediction under the corrupted run and the corrupted-with-restoration run for a specific layer $l$ and position $i$:
$$
\text{IE}_{i}^{(l)} = \mathbb{P}_{\text{corr}, h_i^{(l), \text{clean}}}[q] - \mathbb{P}_{\text{corr}}[q]
$$

The average total effect (ATE) is the mean total effect across multiple samples:
\begin{equation}
\text{ATE} = \frac{1}{N} \sum_{i=1}^{N} \text{TE}_i
\end{equation}

The average indirect effect (AIE) calculates the mean indirect effect for each hidden state variable $h_i^{(l)}$:
\begin{equation}
\text{AIE}^{(l)} = \frac{1}{N} \sum_{i=1}^{N} \text{IE}_i^{(l)}
\end{equation}

By analyzing these effects, we can identify the layers and specific hidden states within the layers that have the most significant causal impact on sentiment polarity predictions.

\subsection{Causal Tracing Results}

\begin{figure}[t]
    \captionsetup{type=figure}
    \centering
    \includegraphics[width=\linewidth]{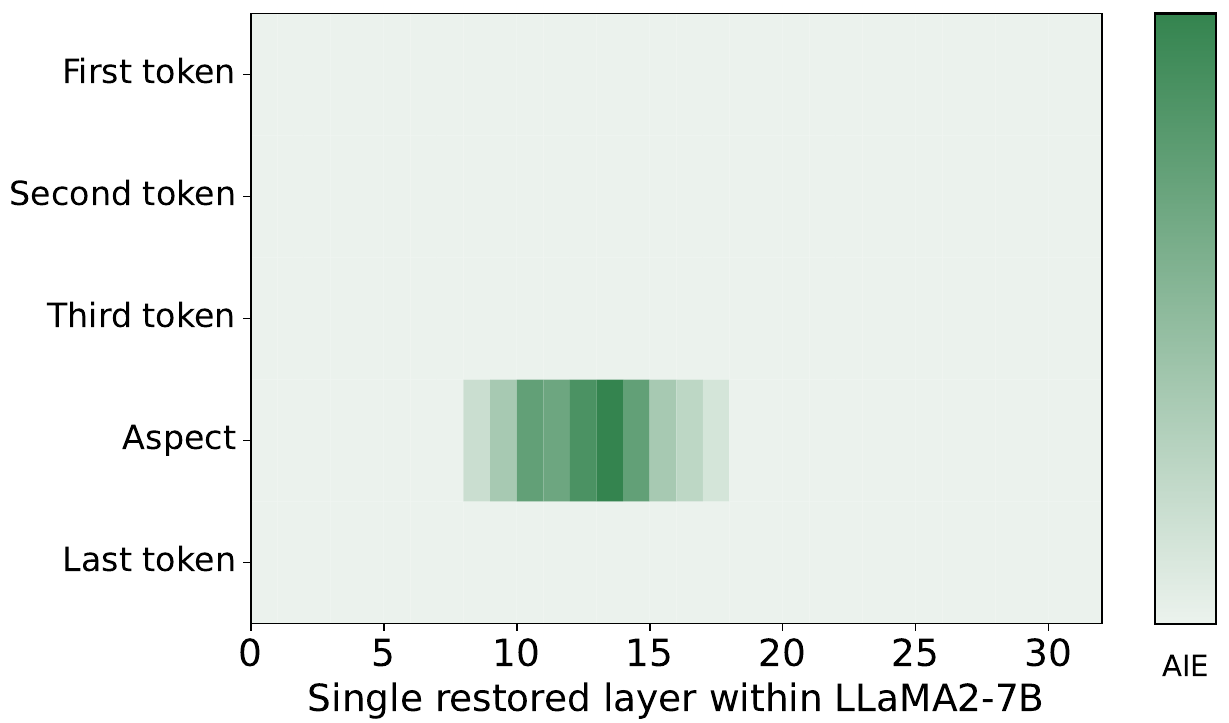}
    \caption{
        \label{fig:causal_results} 
            Causal tracing results for ABSC at various words, layers, and positions.
            The "First token," "Second token," and so on denote the positions of the input sentence tokens. 
    }
\end{figure}

We begin by adapting the LLM\footnote{https://huggingface.co/meta-llama/Llama-2-7b-hf} to solve the ABSC task. 
Subsequently, we compute the average indirect effect over 400 samples from the ABSC dataset.

Since the model inevitably makes errors, we use the ATE to discard outliers to ensure an accurate interpretation of the causal tracing results.
Specifically, we retain only the results that met the expected prediction: $\mathbb{P}[q]$ for correct prediction and $\mathbb{P}_{\text{corr}}[q]$ for incorrect prediction.
 
As illustrated in Figure~\ref{fig:causal_results}, our exploration focuses on the influence of tokens at different positions and layers within the model. 
Our findings reveal that a notable portion of the effect is mediated by highly causal individual states, particularly at the aspect tokens.

While it is expected that aspect tokens play a decisive role in predicting sentiment polarity, the causal tracing results explicitly demonstrate that the mid-layer representations of aspect tokens play a significant role in ABSC.
The emergence of these causal states in the mid-layers sheds light on more efficient parameter-efficient fine-tuning methods for ABSC.

\subsection{Model Editing for LLM-based ABSC}
Given the importance of mid-layer representations of aspect tokens for ABSC, we propose that employing model editing in only these representations can guide the model to solve ABSC with significantly fewer parameters. 
Our method leverages the strengths of both weight-based~\cite{hu2021lora} and representation-based editing~\cite{wu2024reft} to achieve this goal.

Recall that in each layer of the decoder model, the representation is updated through a specific process.
The hidden state $\mathbf{h}_l$ at layer $l$ is derived by incorporating the previous layer's representation $\mathbf{h}_{l-1}$, the output from the attention mechanism, and the output from a multi-layer perceptron (MLP). Formally, this can be expressed as:
\begin{equation}
\mathbf{h}_t = \mathbf{h}_{t-1} + a(\mathbf{h}_{t-1}) + \text{MLP}(\mathbf{h}_{t-1} + a(\mathbf{h}_{t-1}))\text{,}
\end{equation}
where $a(\cdot)$ denotes the attention mechanism and MLP$(\cdot)$ denotes the multi-layer perceptron.

\subsubsection{Weight-based Editing}
Previous works~\cite{TangJLZ20,WangSYQW20,ZhangQ20} have demonstrated that the attention mechanism is crucial for capturing relationships between tokens, especially aspect tokens. 
Therefore, we only employ weight-based model editing on the attention output projection matrix, which is responsible for transforming these multi-head attention outputs back into the model's hidden space, directly affecting how aspect token relationships are encoded. 
By applying weight-based editing specifically to this matrix, we can edit the model to handle aspect tokens with minimal disruption to other layers, ensuring parameter efficiency while maintaining strong performance. 

Specifically, we update the weight matrix $\mathbf{W}_l$ of the targeted layers with weight-based editing~\cite{hu2021lora} using low-rank matrix $\mathbf{A}_l$ and $\mathbf{B}_l$ .

Let $\mathbf{W}_{l} \in \mathbb{R}^{d \times k}$ represent the attention layer output projection matrix of the $l$-th layer in the LLM. We introduce low-rank matrices $\mathbf{A}_l \in \mathbb{R}^{d \times r}$ and $\mathbf{B}_l \in \mathbb{R}^{r \times k}$, such that the weight update can be approximated as:
\begin{equation}
\Delta \mathbf{W}_l \approx \mathbf{A}_l \mathbf{B}_l
\end{equation}

The adapted weight matrix $\mathbf{W}_l'$ is then:

\begin{equation}
\mathbf{W}_l' = \mathbf{W}_l + \Delta \mathbf{W}_l = \mathbf{W}_l + \mathbf{A}_l \mathbf{B}_l
\end{equation}

\subsubsection{Representation-based Editing}
Previous works~\cite{pmlr-v162-ravfogel22a,avitan2024changed,singh2024mimic} demonstrate that editing the residual stream is crucial for controlling pretrained LM outputs without intensive finetuning. 
Therefore, we focus on editing the residual stream representation to activate important components within the parameters of the LLM for ABSC. 

For the representation editing, we apply linear representation editing using projection matrix $\mathbf{R}_l$ and linear interventions $\mathbf{W}_l^*$. Specifically, we define a low-rank projection matrix $\mathbf{R}_l \in \mathbb{R}^{r \times d}$ with orthonormal rows. 
The modified hidden representation $\mathbf{h}_l^t$ at layer $l$ for an aspect term at position $t$ is given by:

\begin{equation}
\alpha(\mathbf{h}_l^t) = \mathbf{h}_l^t + \mathbf{R}_l^\top (\mathbf{W}_l^* \mathbf{h}_l^t + \mathbf{b}_l - \mathbf{R}_l \mathbf{h}_l^t),
\end{equation}
where $\mathbf{W}_l^* \in \mathbb{R}^{r \times d}$ is a linear projection matrix and $\mathbf{b}_l$ is a bias vector.

Thus, as shown in Figure~\ref{fig:flow_chart}(e), the overall optimization objective during model editing for ABSC with our method is to minimize the loss $\mathcal{L}$ with respect to the model editing parameters of $\theta$ = $\{\mathbf{R}_l,\mathbf{W}_l^*,\mathbf{A}_l, \mathbf{B}_l, \mathbf{b}_l\}_{l \in L^*}$, 
 
\begin{equation}
    \mathcal{L}(\theta) = \mathbb{E}_{(x,y) \sim \mathcal{D}} \left[ -\sum_{t=1}^{|y|} \log \left( f_{\Phi_0 + \Delta \Phi(\theta)} (y_t \mid x, y_{<t}) \right)\right],
\end{equation}
where $L^*$ denotes the specific training layers. 

Notably, the task specific model editing parameter $|\theta|$ is much smaller than the LLM parameters $|\Phi_0|$.
Therefore, our method allows the model to be efficiently edited at critical layers and positions, thereby improving performance on ABSC with significantly fewer parameters.

\begin{table}[t]
\centering
\begin{tabular}{lcr}
\toprule
\textbf{Dataset} & \textbf{Split} & \textbf{Number of Reviews} \\
\midrule
Device & Train & 1,394 \\
& Test & 691 \\
\midrule
Laptop & Train & 2,297 \\
& Test & 631 \\
\midrule
Restaurant & Train & 4,284 \\ 
& Test & 2,252 \\
\midrule
Service & Train & 1,840 \\
& Test & 886 \\
\bottomrule
\end{tabular}
\caption{Distribution of reviews across different domains.}
\label{table-dataset}
\end{table}
\begin{table*}[!t]
\centering
{
\begin{tabular}{llccccccc}
    \toprule
	\multirow{2}{*}{\textbf{Methods}}  &\multirow{2}{*}{\textbf{Base Model}} &\multirow{2}{*}{\textbf{Params} (\%)}&\multicolumn{4}{c}{\textbf{Accuracy} ($\uparrow$)} \\ \cmidrule{4-8}
	& & & \textbf{Device}  & \textbf{Laptop} & \textbf{Restaurant} & \textbf{Service}& \textbf{Avg.}\\
    \midrule
    \multicolumn{8}{l}{\it{Zero-Shot}} \\
    GPT4o-mini~\cite{gpt4omni2024} 
    & ---  &---  &87.6   &\textbf{80.0}  &85.2&86.7&84.9 \\ 
    Llama2-7b~\cite{touvron2023llama} 
    & ---  &---  &68.7   &53.2  &66.5&56.0&61.1 \\ 
    \cmidrule{1-8}
    \multicolumn{8}{l}{\it{Full-parameter-fintuning}} \\
    BERT~\cite{bert18}
    & BERT-base &1.632\%  &90.7 &71.2  &84.8 &82.4&82.3\\ 
	
    Deberta~\cite{he2021deberta}
    & DebertaV3-base &2.057\%  &95.4  &74.2 &86.3 &86.0&85.5\\
    
    Flan-T5~\cite{flanT5}
    & Flan-T5-base &3.265\% &94.5  &75.8  &86.5 &87.4&86.1 \\

    \midrule
        
    \multicolumn{8}{l}{\it{Parameter-efficient finetuning}} \\
    Adapter~\cite{Adapter19}
    & Llama2-7b & 1.953\% &73.2   &68.6  &72.1  &75.2&72.3\\
    
    LoRA~\cite{hu2021lora}
    & Llama2-7b & 0.826\% & 95.2  &76.9  &85.7  &88.3 &86.5\\
    
    Dora~\cite{liudora2024}
    & Llama2-7b & 0.838\% & 95.1  &77.2  &85.9  &88.1&86.6 \\

    PrefixFT~\cite{li2021prefix}
    & Llama2-7b & 0.039\% &68.9   &53.2  &66.5  &37.6&56.6\\ 
    
    LoReft~\cite{wu2024reft}
    & Llama2-7b & 0.031\% & 94.2 & 73.7 & 87.3 &88.5&85.9 \\
    \midrule
    \multicolumn{8}{l}{\it{Specific-layers finetuning}} \\
    LoRA~\cite{hu2021lora}
    & Llama2-7b & 0.129\% & 93.6 & 73.2 &85.5 &86.8 &84.8\\
    LoReft~\cite{wu2024reft}
    & Llama2-7b & 0.006\% & 92.9 & 72.3  &86.9 &87.6 &84.9\\
    \textbf{Ours}~
    & Llama2-7b & 0.006\% & \textbf{96.2} & 76.7 &\textbf{88.1} &\textbf{89.5} &\textbf{87.6}\\
    \bottomrule
\end{tabular}
}
\caption{
        \label{tab:in-domain-results}
        Overall accuracy (\%) over four in-domain aspect-based sentiment classification datasets. The "Params" column indicates the proportion of trainable parameters relative to the number of parameters in Llama2-7b. Specific-layers finetuning refers to updating only the middle layers, specifically layers 10 to 15.
    }
\end{table*}

\section{Experiment}
In this section, we introduce our experimental setup and implementation details, present our method's performance on in-domain and out-of-domain ABSC tasks compared to competitive baselines, and empirically analyze the effectiveness of our method.
\subsection{Setup}
The labeled dataset used in our experiments includes reviews from four different domains: Restaurant (R), Laptop (L), Device (D), and Service (S). 
Restaurant (R) is a combination of the restaurant reviews from SemEval 2014/2015/2016 ~\cite{PontikiGPPAM14,PontikiGPMA15,PontikiGPAMAAZQ16}. 
Laptop (L) is sourced from SemEval 2014~\cite{PontikiGPPAM14}. 
Device (D) consists of all the digital device reviews collected by~\citet{ToprakJG10}. Service (S) contains reviews from web services introduced by~\citet{HuL04}. 
The distribution of reviews in these domains is detailed in Table~\ref{table-dataset}.

We employ Llama-2-7b~\cite{touvron2023llama} as our primary base large language model. AdamW~\cite{loshchilov2018decoupled} is used as the optimizer, with a learning rate of $3 \times 10^{-4}$ for the low-rank weight projection part and $1 \times 10^{-5}$ for the representation editing part. For the comparison methods, we adopt standard experimental settings and commonly used parameters. 
Specifically, LoRA and Dora utilize a learning rate of $1 \times 10^{-4}$ with rank of 32.  Additionally, we include LoReft with a learning rate of $2 \times 10^{-5}$ with rank of 8.
All comparison experiments are conducted on a single NVIDIA 3090 GPU and we take accuracy as the evaluation metric. 
The experimental results are obtained by averaging three runs with random initialization.
The PEFT methods are trained for one epoch, while the full parameter methods are trained for three epochs.

The causal tracing results reveal that the representations within mid-layers, specifically layers 10-15, show a decisive influence on ABSC. 
Therefore, in the \textit{specific-layer finetuning} section of our experimental results, all methods are restricted to these specific layers. 
Instead, in the section of parameter-efficient finetuning, these methods are trained on all layers.

\subsection{Main Results}
To assess the effectiveness of the proposed method, we mainly conduct two experiments: in-domain and out-of-domain evaluations.
We compare our method with zero-shot methods~(i.e., GPT-4o-mini, Llama2-7b), full-parameter finetuning methods~(i.e., BERT, Deberta, Flan-T5), and parameter-efficient finetuning methods~(i.e., PrefixFT, Adapter, LoRA, Dora, Loreft).

\subsubsection{In-domain Analysis}

In-domain ABSC tasks refer to the task that the training data and testing data are in the same domain. This allows us to assess how well the methods perform on data similar to what they were trained on. 

As shown in Table~\ref{tab:in-domain-results}, our methods achieve the highest average accuracy, significantly outperforming other methods with fewer trainable parameters. 
Llama2-7b with zero-shot manner exhibits significantly lower performance compared to other method, indicating its limited effectiveness in the absence of domain-specific fine-tuning.
The full-parameter finetuning models show improved performance. This suggests that full-parameter finetuning can effectively enhance model performance by leveraging domain-specific data.
Among the PEFT methods, PrefixFT, despite its small tuning parameter size, shows only marginal improvement over zero-shot Llama2-7b. 
LoRA and Dora exhibit considerably better performance. However, these methods still involve larger tuning parameter sizes compared to our proposed methods.
LoReft, while efficient in terms of parameter size, does not perform as well as our methods.
These results indicate our method not only achieves high accuracy but also exhibits remarkable parameter efficiency.

\subsubsection{Out-of-domain Analysis}
Due to the high risk of catastrophic forgetting in LLM-based in-domain training, especially in PEFT methods, we design out-of-domain analysis to evaluate the ability of methods to generalize on out-of-distribution data.

As shown in Table~\ref{tab:out-of-domains}, our approach significantly outperforms other strong baselines while maintaining a lower number of trainable parameters. 
Although we are concerned about the potential risk of data leakage with GPT4o-mini~\cite{Koco2023}, our method still surpasses GPT-4o-mini in some domains and in average accuracy.
This indicates the effectiveness of our method in handling data variability and ensuring robust generalization across different domains.
LoRA and LoReft, as state-of-the-art PEFT methods, show strong performance in specific-layers finetuning, but these methods are still vulnerable in some out-of-distribution samples.
Our methods, based on the causal tracing results of sentiment association, not only achieve high accuracy but also maintain computational efficiency. 
The results clearly indicate that our method offers a balanced combination of accuracy and efficiency.

\begin{table*}[!t]
\centering
{
\small
\begin{tabular}{llccccccccccc}
    \toprule
	\multirow{2}{*}{\textbf{Methods}} & \multicolumn{11}{c}{\textbf{Accuracy} ($\uparrow$)} \\ \cmidrule{2-12}
	& \textbf{(D, R)} & \textbf{(D, S)} & \textbf{(L, R)} & \textbf{(L, S)} & \textbf{(R, D)} & \textbf{(R, L)} & \textbf{(R, S)} & \textbf{(S, D)} & \textbf{(S, L)} & \textbf{(S, R)} & \textbf{Avg.} \\
    \midrule
    \multicolumn{12}{l}{\it{Zero-Shot}} \\
    GPT-4o-mini~\cite{gpt4omni2024} 
    &85.2  &86.7  &\textbf{85.2}  &\textbf{86.7}  &87.6  &80.0  &86.7  &87.6  &\textbf{80.0}  &\textbf{85.2} &85.1 \\ 
    Llama2-7b~\cite{touvron2023llama}
    &66.5  &56.0  &66.5  &56.0  &68.7  &53.2  &56.0  &68.7  &53.2  &66.5  &61.1  \\ 
    
    \cmidrule{1-12}
    \multicolumn{12}{l}{\it{Full-parameter-fintuning}} \\
    BERT~\cite{bert18}
    & 80.4     & 85.2   & 81.1     & 81.9     & 88.3     & 77.8     & 84.5     & 90.7     & 69.7     & 79.8     &81.9  \\ 
	
    Deberta~\cite{he2021deberta}
    & 81.1     & 85.4   & 80.1     & 82.7     & 88.3     & 78.4     & 86.2     & 93.9     & 68.5     & 81.0     &82.6\\
    
    Flan-T5~\cite{flanT5}
    & 81.8     & 87.6    & 79.5     & 85.3     & 93.9   & 78.3     & 87.0     & 93.5 & 70.4   & 81.2     &83.9  \\

    \midrule
        
    \multicolumn{12}{l}{\it{Parameter-efficient finetuning}} \\
    LoRA~\cite{hu2021lora}
    &82.9  &87.2  &85.0  &81.6  &84.9  &79.2  &75.2  &94.4  &70.7  &83.8  &82.5 \\
    
    Dora~\cite{liudora2024}
    &82.4  &87.4  &84.7  &81.2  &84.5 &\textbf{80.2}  &75.1  &94.6  &69.9  &84.1  &82.4 \\

    LoReft~\cite{wu2024reft}
    &84.0  &87.6  &83.2  &85.2  &94.5  &72.9  &\textbf{88.5}  &94.1  &71.3  &84.4  &84.6 \\
    
    \midrule
    
    \multicolumn{12}{l}{\it{Specific-layers finetuning}} \\
    LoRA~\cite{hu2021lora} 
    &81.9  &86.2  &84.2  &80.1  &83.2  &77.5  &72.3 &92.8  &67.2  &82.1   &80.8 \\
    LoReft~\cite{wu2024reft}
    &83.8  &83.4  &83.3  &83.3  &\textbf{95.2}  &72.4  &86.6  &92.2  &65.9  &83.3  &82.9 \\
    \textbf{Ours}~
    &\textbf{85.8}  &\textbf{88.7}  &85.0  &85.2  &92.9  &75.6  &\textbf{88.5}  &\textbf{95.8}  &71.6  &84.9  &\textbf{85.4} \\
    \bottomrule
\end{tabular}
}
\caption{
        \label{tab:out-of-domains}
        Overall accuracy (\%) over 12 out-of-domain aspect-based sentiment classification dataset pairs. The column "(D,R)" indicates the model is trained on \textbf{D}evice domain and tested on \textbf{R}estaurant domain, except these zero-shot methods. Specific-layers finetuning refers to updating only the middle layers, specifically layers 10 to 15.
    }
\end{table*}

\subsection{Impact on Specific Layers}
\begin{figure}[t]
    \captionsetup{type=figure}
    \centering
    \includegraphics[width=\linewidth]{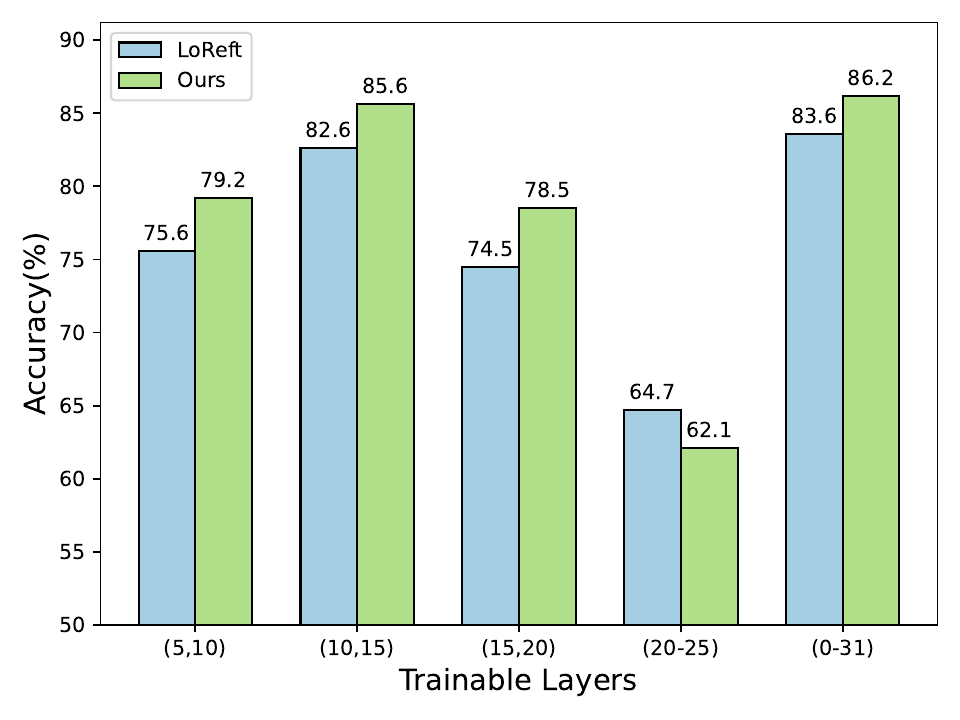}
    \caption{
        \label{fig:influence-layers} 
            Impact on trainable layers of LoReft and our method. 
    }
\end{figure}

As the causal tracing results mentioned before, the mid-layers show great influence on predicting the sentiment polarity of the given aspect. 
To further explore this impact, we conduct a quantitative experiment to analyze its impact on the performance of different methods in average accuracy over 16 domain pairs, including in-domain and out-of-domain datasets.

As shown in Figure~\ref{fig:influence-layers}, different methods exhibit consistent results across various layers. This consistency indicates the significant influence of different layers in the LLM on addressing the ABSC task. Specifically, within the 10-15 layer range, both methods achieve results comparable to their best performance when training parameters across all layers (0-31). 
This finding aligns with our previously mentioned causal tracing sentiment association results.
Moreover, our method demonstrates a more stable and superior performance compared to LoReft. 
This stability is particularly notable across all evaluated layer ranges, further indicating the robustness and effectiveness of our method.

\subsection{Influence of Specific Positions}
\begin{figure}[t]
    \captionsetup{type=figure}
    \centering
    \includegraphics[width=\linewidth]{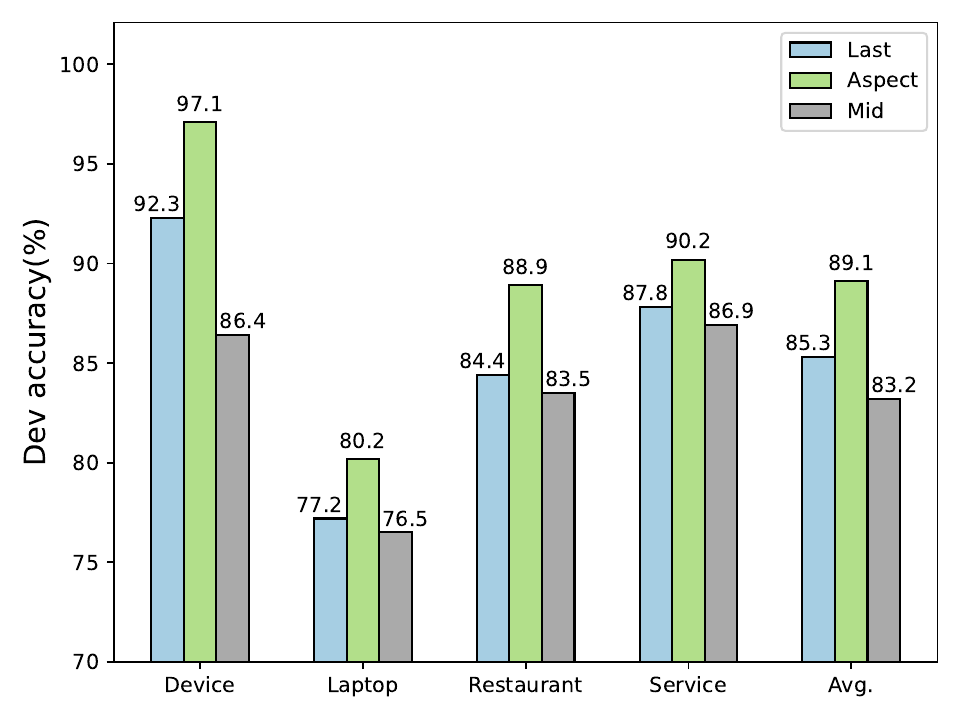}
    \caption{
        \label{fig:influence-positions} 
            Influence on specific positions: 'aspect' refers to editing on aspect terms, 'last' refers to editing on the last words, and 'mid' refers to editing on a random mid-position word.
    }
\end{figure}

For our method, we must decide which layers and input positions to apply the intervention on. 
The causal tracing results reveal that aspect terms in mid-layers have a significant influence on the sentiment prediction. 
In this section, we investigate three potential word positions within the specific layers (10-15) to evaluate our methods.
Intuitively, the last word and aspect word may have a strong impact, but we also take a random mid-word from the input to evaluate the influence of the intervention position. 
In this experiment, we take 10\% of data from the training dataset as the development dataset for in-domain ABSC.

As shown in Figure~\ref{fig:influence-positions}, editing on aspect terms yields the highest performance. It highlights the pivotal role aspect terms play in ABSC.
Additionally, the last word shows relatively good performance due to the use of a decoder-only model. However, the performance still falls short compared to editing on aspect terms. 
This indicates that for specific tasks, targeting the key terms relevant to the task, such as aspect terms, can lead to better results.
Notably, the performance does not degrade significantly even when a random mid-position word is chosen for intervention. It suggests that while targeted editing on aspect terms is optimal, our method, which containing weight,and representation based editing methods, remains robust and does not suffer from significant performance drops due to random selection.

\section{Conclusion}
We explore model editing as an efficient and interpretable fine-tuning method for LLM-based aspect-based sentiment classification. 
By combining causal tracing with targeted model edits, we demonstrate that model editing can serve as an efficient fine-tuning method with minimal parameter updates. 
In addition, we empirically validate that specific mid-layer representations of aspect words play a crucial role in aspect-based sentiment classification.
Practical experiments confirm the effectiveness of our framework, demonstrating that it outperforms state-of-the-art parameter-efficient fine-tuning methods with fewer trainable parameters. 
These results indicate the potential of model editing as a more interpretable and efficient strategy for adapting LLMs to specific downstream tasks, highlighting its broader applicability.

\section{Acknowledgments}
We would like to thank Yanzhi Xu, Tianlai Ma and the anonymous reviewers for their insightful and valuable comments. 
This work is partially supported by the National Natural Science Foundation of China (No. 62276177) and Project Funded by the Priority Academic Program Development of Jiangsu Higher Education Institutions.

\bibliography{aaai25}

\clearpage

\end{document}